\documentclass[letterpaper, 10 pt, conference]{ieeeconf}  

\IEEEoverridecommandlockouts                              

\usepackage{graphics} 
\usepackage{epsfig} 
\usepackage{mathptmx} 
\usepackage{times} 
\usepackage{amsmath} 
\usepackage{amssymb}  
\usepackage{xcolor}  
\usepackage{soul}   

\usepackage[pagebackref=true,breaklinks=true,letterpaper=true,colorlinks,bookmarks=false]{hyperref}
\usepackage{multirow}
\usepackage{multicol}
\usepackage{mathtools}
\usepackage{graphicx}
\usepackage{caption}
\usepackage{array}
\usepackage{cite}
\usepackage{bbding}
\usepackage{diagbox}
\title{\LARGE \bf
You Only Scan Once: A Dynamic Scene Reconstruction Pipeline for 6-DoF Robotic Grasping of Novel Objects
}

\author{Lei Zhou$^{1,*}$, Haozhe Wang$^{1,2}$, Zhengshen Zhang$^{1}$, Zhiyang Liu$^{1}$, Francis EH Tay$^{1}$, and Marcelo H. Ang Jr$^{1}$
\thanks{$^{*}$Corresponding Author.}
\thanks{$^{1}$Authors are with the Advanced Robotics Centre, National University of Singapore, 117608, Singapore. {\tt\small \{leizhou, wang\_haozhe, zhengshen\_zhang, zhiyang\}@u.nus.edu}, {\tt\small \{mpetayeh, mpeangh\}@nus.edu.sg}}%
\thanks{$^{2}$Haozhe Wang is with the Integrative Sciences and Engineering Programme, National University of Singapore Graduate School, 119077, Singapore.}%
}

\begin{document}

\maketitle
\thispagestyle{empty}
\pagestyle{empty}

\begin{abstract}

In the realm of robotic grasping, achieving accurate and reliable interactions with the environment is a pivotal challenge. Traditional methods of grasp planning methods utilizing partial point clouds derived from depth image often suffer from reduced scene understanding due to occlusion, ultimately impeding their grasping accuracy. Furthermore, scene reconstruction methods have primarily relied upon static techniques, which are susceptible to environment change during manipulation process limits their efficacy in real-time grasping tasks. To address these limitations, this paper introduces a novel two-stage pipeline for dynamic scene reconstruction. In the first stage, our approach takes scene scanning as input to register each target object with mesh reconstruction and novel object pose tracking. In the second stage, pose tracking is still performed to provide object poses in real-time, enabling our approach to transform the reconstructed object point clouds back into the scene.
Unlike conventional methodologies, which rely on static scene snapshots, our method continuously captures the evolving scene geometry, resulting in a comprehensive and up-to-date point cloud representation. By circumventing the constraints posed by occlusion, our method enhances the overall grasp planning process and empowers state-of-the-art 6-DoF robotic grasping algorithms to exhibit markedly improved accuracy.

\end{abstract}

\section{INTRODUCTION}
\label{sec:I}
\begin{figure}[!t]
	\centering
	\includegraphics[width=\linewidth]{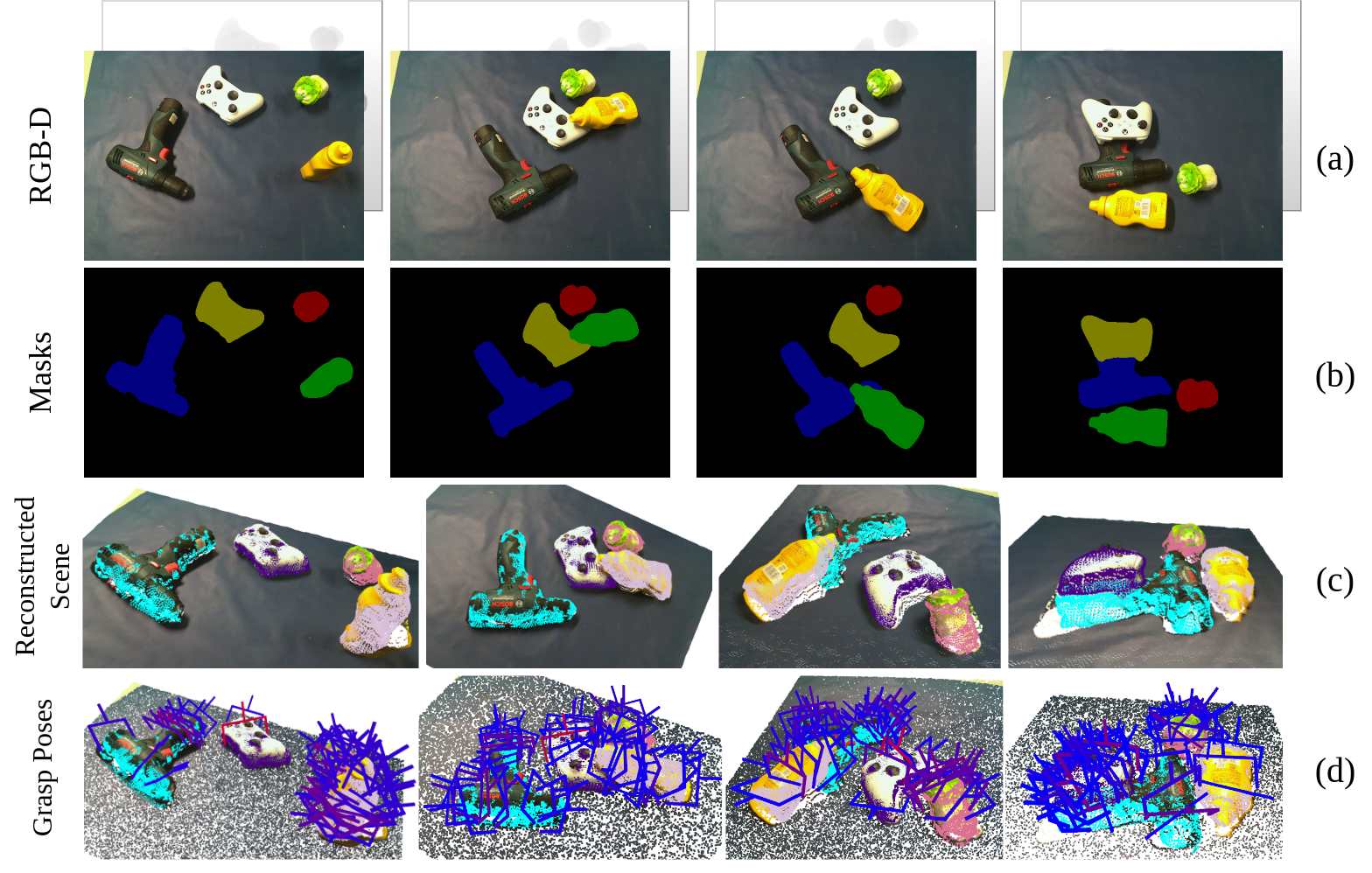}
	\caption{Dynamic scene reconstruction and grasp generation. (a) RGB-D images are captured by an RGB-D camera as it scans the grasping workspace. (b) A \textit{Video-segmentation Module} segments the graspable objects in the scene. (c) Using the RGB-D images and masks from (a) and (b), we reconstruct the meshes of the graspable objects and merge them with the original partial point cloud to create a full point cloud of the workspace. (d) Finally, a \textit{Grasp Pose Predictor} is used to generate the valid grasps based on the reconstructed full point cloud.}
	\label{teaser}
\vspace{-6mm}
\end{figure}

In the realm of robotic manipulation, the ability to grasp and interact with objects in dynamic and complex environments remains a cornerstone challenge. Achieving effective grasping hinges on the fusion of accurate scene understanding and real-time adaptability, which has traditionally posed significant hurdles for existing methodologies \cite{Duan2021survey,Newbury2022survey}. Previous efforts in grasp planning and scene reconstruction have primarily gravitated towards either partial point cloud utilization or static scene representations, each marked by inherent limitations in capturing the dynamic nature of real-world scenarios. Robotic tasks involving some form of 3D
visual perception greatly benefit from a complete knowledge
of the working environment.

Grasp generation methods \cite{fang2020graspnet,Ma_2022_CoRL,liang2019pointnetgpd,Ni2020PointNetGL,wei2021gpr}, reliant on partial point clouds back-projected from depth image, introduce their own set of constraints. The quality of grasp generation directly hinges upon the accuracy and completeness of the partial views captured. Owing to occlusions and partial observability \cite{Sundermeyer2021contactgraspnet}, the objects (or parts) suitable for grasping are invisible from a single viewpoint. The resultant grasp plans lack a comprehensive understanding of the scene, subsequently compromising both accuracy and diversity of robotic grasping.

By predicting the missing part of the object point cloud, the full shape of an object can be recovered to generate more diverse grasps \cite{lundell2020beyond,mohammadi20233dsgrasp,hidalgo2023anthropomorphic}. However, these methods introduce uncertainty to the generated grasps at the same time as the generated points are unreliable compared to the initially observed point cloud. 

By leveraging multi-view input, static scene reconstruction methods, exemplified by NeRF-based \cite{IchnowskiAvigal2021DexNeRF,kerr2023evo-nerf,Dai2023GraspNeRF} and TSDF-based \cite{breyer2020volumetric, jiang2021synergies,Cai2022VolumetricbasedCP} approaches, have demonstrated commendable efficacy in reconstructing environments to achieve more accurate and diverse grasp generation. However, these methods rely on static snapshots, inherently incapable of adapting to changes after scanning. Consequently, though the marriage of such methods with robotic grasping tasks has demonstrated the ability to generate more diverse grasps compared to taking partial point cloud as input, its applicability is impeded in robotic manipulation environments marked by constant change.


To bridge these disparate domains and unlock a new paradigm in robotic grasping, we present \textit{You Only Scan Once} (YOSO), a novel approach that harmonizes the benefits of both static scene reconstruction and partial point cloud-based grasp planning. As shown in Fig. \ref{teaser}, our proposed pipeline introduces a dynamic scene reconstruction methodology that operates in real-time to complete the object point cloud in the scene for subsequent robotic grasping tasks. Unlike conventional static methods that need to repetitively scan the scene when it is changed, our approach only scans the scene once to generate mesh for each novel object in the scene. After that, it dynamically tracks the object pose and transforms the generated object mesh back into the scene to encompass the evolving environment.



Through comprehensive testing, we have assessed the effectiveness of our approach. Our results indicate the substantial improvement in grasping accuracy achieved by providing more complete scene understanding for the grasp planning process while operating at near real-time. 
Our contributions can be summarized as follows:
\begin{enumerate}
    \item We propose a novel and modularized pipeline, YOSO, for dynamic scene reconstruction tailored to the context of robotic grasping tasks.
    \item  We evaluate a pre-trained  state-of-the-art (SOTA) grasp generation model on our reconstructed scene and demonstrate that replacing the input partial point cloud with a more informative reconstructed scene from YOSO pipeline enables it to surpass its current SOTA evaluation results on the GraspNet-1Billion benchmark.
    \item We also extend the GraspNet-1Billion dataset to include the completed point cloud of each scene. This addition aims to establish a theoretical upper limit of performance for models when provided with a fully visible point cloud of a scene within the dataset.
\end{enumerate}

\section{RELATED WORKS}
\label{sec:II}

\subsection{Grasping Methods Utilizing Partial Point Clouds}
In recent years, the field of robotic grasping has witnessed significant advancements in leveraging partial point clouds as a critical sensory input for grasp planning. Several notable works have contributed to the exploration of this area, each with its unique approach and methodologies. Fang \textit{et al.} proposed both a popular benchmark dataset for general object grasping consisting of partial point cloud scenes with more than 1 billion annotated grasps, as well as a baseline model to generate 6-Degree-of-Freedom (6-DoF) grasps from partial point clouds~\cite{fang2020graspnet}. Wang \textit{et al.} improved upon the baseline by adding a graspness model which utilizes geometry cues to distinguish graspable areas in cluttered scenes~\cite{wang2021graspness}. Sundermeyer \textit{et al.} proposed Contact-GraspNet~\cite{Sundermeyer2021contactgraspnet}, which treats the 3D points of a partial point cloud as potential grasp contacts. By rooting the full 6-DoF grasp pose and width in the observed point cloud, the dimensionality of the grasp representation can be reduced to 4-DoF, which greatly facilitates the learning process. Ma and Huang~\cite{Ma_2022_CoRL} proposed a Scale Balanced Learning loss and an Object Balanced Sampling strategy to address the challenge of generating accurate grasp poses for small-scale samples. However, the reliance on partial point clouds in robotic grasping still faces significant challenges. The primary issue is the incomplete data these point clouds provide, capturing only visible object surfaces. This limitation can obscure vital details about an object's shape and orientation and potentially lead to suboptimal grasps.



\subsection{Grasping Methods Utilizing Single-view Shape Completion}

In order to obtain a more complete and informative geometric understanding of the scene, some previous works took partial point cloud as input to predict the missing parts of objects \cite{yu2021pointr, wang2021voxel}. 
Lundell \textit{et al.} represented objects as voxels and trained a deep learning network for object completion, along with Carlo (MC) sampling enhanced by dropout techniques \cite{lundell2019robust}. Subsequently, grasps are predicted on the completed object shapes. However, this voxel-based approach significantly increases memory usage and computational time. Recent efforts \cite{rosasco2022towards,mohammadi20233dsgrasp} have shifted towards using point cloud representations to streamline shape completion and grasp generation processes, aiming for more time-efficient testing phases. Although employing implicit object representations has been shown to accelerate shape completion to 0.7 seconds per object, as demonstrated by \cite{humt2023combining}, such speed remains unsatisfactory for real-time applications, particularly in environments with multiple objects. 


Additionally, the efficacy of those shape completion methods is hindered by their limited ability to generalize to novel object categories not covered in the training data, often leading to less precise reconstructions. In contrast, our proposed pipeline focuses on the reconstruction of novel object shapes, advocating for the use of multi-view inputs over single-view input. This approach not only enriches data reliability but also ensures a more accurate and comprehensive understanding of the scene.

\subsection{Static Scene Reconstruction Methods}

\subsubsection{TSDF-based Methods}
The field of robotic grasping has also witnessed significant advancements through the utilization of Truncated Signed Distance Field (TSDF) representations in scene reconstruction. TSDF is a volumetric approach that represents the scene as a voxel grid. Each voxel stores the distance to the nearest surface. Several works have leveraged this simple and efficient technique to enhance 6-DoF grasp generation and overcome various challenges encountered in robotic manipulation tasks~\cite{breyer2020volumetric, jiang2021synergies, Dai2023GraspNeRF}.

While TSDF-based methods have undeniably advanced the field of robotic grasping, it is imperative to recognize their limitations. Notably, TSDF-based techniques can be computationally expensive, particularly when dealing with large-scale scenes, due to memory and computation requirements for updating the TSDF representation. 



\subsubsection{NeRF-based Methods}

Neural Radiance Field (NeRF)-based methods for static scene reconstruction \cite{mildenhall2020nerf, muller2022instant} has shown potential in capturing and rendering intricate 3D scenes from different perspectives. These approaches are valued for their ability to produce high-quality reconstructions, proving beneficial in areas such as robotic grasping \cite{wang2023real2sim2real}. NeRFs are particularly adept at handling complex scenarios, such as those involving non-Lambertian materials and challenging lighting \cite{IchnowskiAvigal2021DexNeRF,kerr2023evo-nerf,Dai2023GraspNeRF}, offering impressive visual quality in the representations they create.

However, it's important to recognize the limitations associated with NeRF-based reconstruction methods, especially when considering their application in dynamic tasks like 6-DoF robotic grasping. Similar to TSDF-based methods, these NeRF-based methods primarily accommodate static scenes, which limits their direct applicability in environments subject to change, such as those involving moving objects. In an effort to bridge this gap, continual NeRF training, as implemented in Evo-NeRF \cite{kerr2023evo-nerf}, has been developed to facilitate rapid updates to the NeRF model shortly after each grasp. Nevertheless, this process necessitates the re-scanning of the workspace and the acquisition of a small set of images to support continual NeRF training, still assuming that the environment remains unaltered after scanning. Such a requirement makes it less practical in dynamic scenes.

\section{METHOD}

\begin{figure*}[h!]
	\centering
	\includegraphics[width=\linewidth]{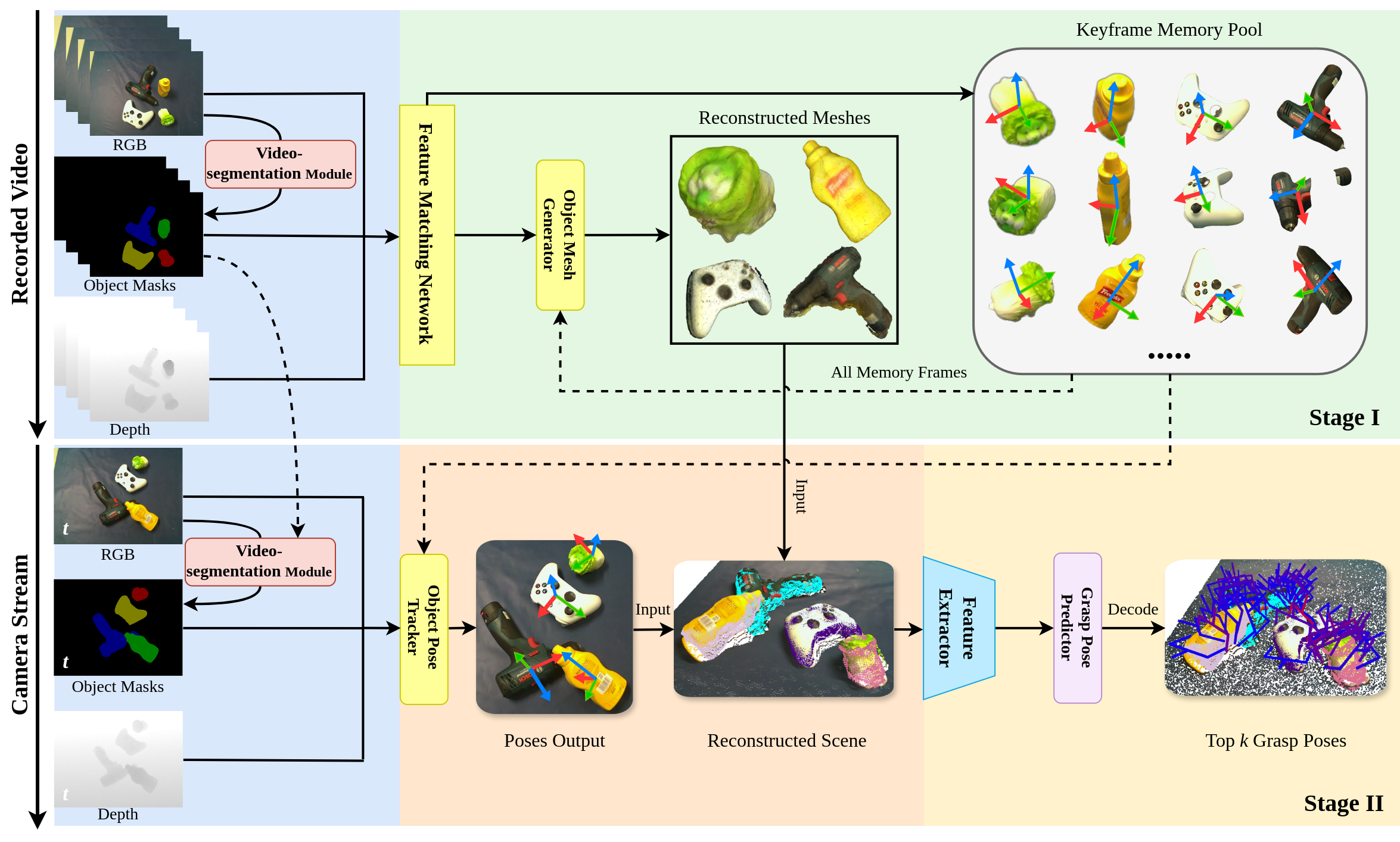}
	\caption{Overview of the proposed pipeline. \textbf{Stage I}:
 Given a monocular RGB-D video, object masks are segmented using a \textit{Video-segmentation Module}. Subsequently, feature matching is performed in the \textit{Object Pose Tracker and Mesh Generator} module to simultaneously track object pose and reconstruct object mesh. Keyframes with informative historical observations are stored in the memory pool to facilitate pose tracking in both stages. \textbf{Stage II}:  In testing, given an RGB-D image, the masks of the objects in the workspace are segmented out and the object pose is estimated by taking the \textit{Keyframe Memory Pool} as a reference. Subsequently, the reconstructed meshes are transformed into camera coordinates with the estimated object pose. Taking this reconstructed scene point cloud, grasp generation is performed to generate the top \textit{k} grasp poses for real-world experiments. The dotted lines represent the supplementation of historical information.}
	\label{pipeline}	
\vspace{-4mm}
\end{figure*}

The proposed pipeline is structured into two distinct stages, denoted as \textbf{Stage I} and \textbf{Stage II} in Fig. \ref{pipeline}. Unlike conventional static scene reconstruction methods that necessitate repetitive re-scanning when changes occur, our approach only perform a single scene scan to register for novel objects in \textbf{Stage I}, subsequently enabling dynamic scene reconstruction in \textbf{Stage II}. Hence, we have coined the term \textit{"You Only Scan Once"} to encapsulate the essence of our methodology.

In \textbf{Stage I}, a camera mounted to the robotic manipulator's wrist scans the scene along a predefined motion trajectory, encompassing the hemispherical region above the workspace. This scanning procedure yields a monocular RGB-D video, referred to as the reference video. Each object within this reference video is registered within a memory pool, wherein their respective poses and images are stored. Concurrently, a separate parallel thread generates the object meshes.

Transitioning to \textbf{Stage II}, we strategically pause the computationally intensive mesh reconstruction process while maintaining real-time tracking of object pose changes relative to the initial frame of the reference video. This adjustment enables the remainder of our pipeline to operate nearly in real-time. By transforming the generated object meshes into camera coordinates, accounting for their 6D poses, we effectively address occluded regions and complete missing object parts. Consequently, we reconstruct a point cloud representation of the scene, subsequently employed as input for the grasp pose prediction network to estimate the 6-DoF pose of the gripper.

Furthermore, our pipeline has been meticulously designed in a modular fashion, comprising three primary components: \textit{Video-segmentation Module} (\ref{sec:IIIA}), \textit{Object Pose Tracker and Mesh Generator} (\ref{sec:IIIB}), and \textit{Grasp Pose Predictor} (\ref{sec:IIIC}). This modular design approach empowers us to enhance overall performance through the integration of advanced algorithms within each module.

It's noteworthy that this pipeline relies on only two key assumptions. The first assumption is that objects should be visible in the initial few frames, acknowledging the constraints of the algorithms incorporated in Section \ref{sec:IIIA} and Section \ref{sec:IIIB}. The second assumption pertains to having access to the segmentation mask of target objects in the initial frame. Beyond these considerations, no additional information is necessitated for the pipeline's operation.

\subsection{Video-segmentation Module}
\label{sec:IIIA}

To prepare masks denoted as $\textbf{\textit{M}}^{t}$ corresponding to objects in each frame of the video $\textbf{\textit{I}}^{t}$ for the two subsequent modules, we employ the unified long-term video object segmentation framework, XMem~\cite{cheng2022xmem}. In \textbf{Stage I}, we utilize the camera to scan the entire scene and capture a reference video. This reference video is then input into the XMem model, which subsequently generates masks corresponding to the target objects for each frame. In \textbf{Stage II}, we process the image of each frame captured by the camera through the XMem model for real-time segmentation, enabling us to track the masks of the target objects. It's important to note that in \textbf{Stage II}, we also provide the XMem model with the segmented masks obtained in \textbf{Stage I}. This additional input enables the model to establish long-term memory, enhancing its segmentation performance.


\subsection{6D Object Pose Tracker and Mesh Generator}
\label{sec:IIIB}

In pursuit of our objective, which involves generating meshes for novel objects and seamlessly integrating them into the scene to complete missing parts, we incorporate the approach outlined in BundleSDF \cite{wen2023bundlesdf}. This approach allows for the simultaneous tracking of object poses and the generation of object meshes aligned with the initial camera pose. Consequently, the object pose $\xi_{i}^{t}$ in frame $t$ can be considered as the object's pose within the camera coordinates.

During \textbf{Stage I}, the RGB-D images $\textbf{\textit{I}}^{t}$ and their corresponding segmentation masks $\textbf{\textit{M}}^{t}$, obtained from the \textit{Video-segmentation Module}, are combined to form an input frame $\textbf{\textit{F}}^{t} = [\textbf{\textit{I}}^{t}, \textbf{\textit{M}}^{t}]$. The primary objective in this stage is to reconstruct the mesh $\textit{O}_{i}$ for each object while ensuring alignment with the camera pose from the initial frame.

In \textbf{Stage II}, a pose tracker is employed to estimate the relative pose change $\Delta\xi_{i}^{t}$ of each object with respect to the pose $\xi_{i}^{0}$ in the initial frame. Notably, this initial frame's pose also serves as the object's pose in the camera coordinates.



\subsubsection{6D Pose Tracker for Novel Object}

Building on the work of Wen \textit{et al.} in \cite{wen2023bundlesdf}, our approach includes feature matching in RGB between the current frame $\textbf{\textit{F}}^{t}$ and the previous frame $\textbf{\textit{F}}^{t-1}$. To facilitate long-term pose tracking, $\textbf{\textit{F}}^{t}$ is designated as a keyframe if the relative feature difference between the two frames surpasses a predefined threshold. These keyframes are subsequently stored in a \textit{Keyframe Memory Pool}. Then, for each new frame, a comparison is made with the nearest $ K $ keyframes in the memory pool. This process facilitates online pose graph optimization, which further refines the pose estimation results and mitigates pose drift.

\subsubsection{NeRF-based Mesh Generator}
Following \cite{wen2023bundlesdf}, a \textit{NeRF-based Mesh Generator} is employed to train an object-centric neural signed distance field (SDF). This SDF learns both the multi-view consistent 3D shape and appearance of the object. Given that the mesh generation process is relatively computationally intensive, it is strategically frozen during testing to ensure the real-time performance of our pipeline.

\subsection{6-DoF Grasp Pose Predictor}
\label{sec:IIIC}
In \textbf{Stage II}, given a partial point cloud $\textit{\textbf{P}}^{t}$ back-projected from depth image , the reconstructed object point cloud is first transformed into the camera coordinates based on the tracked poses:
\begin{equation}
    \textit{O}_{i}^{t} = \Delta\xi_{i}^{t}\textit{O}_{i}.
\end{equation}
Subsequently, the scene is reconstructed by merging the observed scene point cloud and the transformed object point cloud: 
\begin{equation}
    \textbf{\textit{P}}^{t}_{merge} = [\textbf{\textit{P}}^{t}, \textbf{\textit{O}}^{t}],
\end{equation} 
where $\textbf{\textit{O}}^{t} = [\textit{O}^{t}_{1}, \textit{O}^{t}_{1}...,\textit{O}^{t}_{M}]$ is the reconstructed point clouds of all target objects and $M$ represents number of objects.

In our pipeline, we incorporate Scale-balanced GraspNet \cite{Ma_2022_CoRL} as our baseline, which is one of the SOTA 6-DoF grasp pose prediction networks. Taking point cloud and segmentation mask from the previous module as input, it predicts a set of grasp poses $\textbf{G}$, which can be further interpreted as the 6-DoF pose of a gripper.

\textit{Grasp Pose Predictor} aims to predict the orientation and translation of the gripper under the camera coordinates, as well as the width of the gripper. We represent the grasp pose $\mathbf{G}$ following \cite{fang2020graspnet} as:
\begin{equation}
    \mathbf{G}=[\mathbf{R}, \quad \mathbf{t}, \quad w],
\end{equation}
where $\mathbf{R} \in \mathbb{R}^{3 \times 3}$ denotes the gripper orientation, $\mathbf{t} \in \mathbb{R}^{3 \times 1} $ denotes the center of grasp and $w \in \mathbb{R}$ denotes the gripper
width that is suitable for grasping the target object.
\begin{figure}[t]
	\centering
	\includegraphics[width=\linewidth]{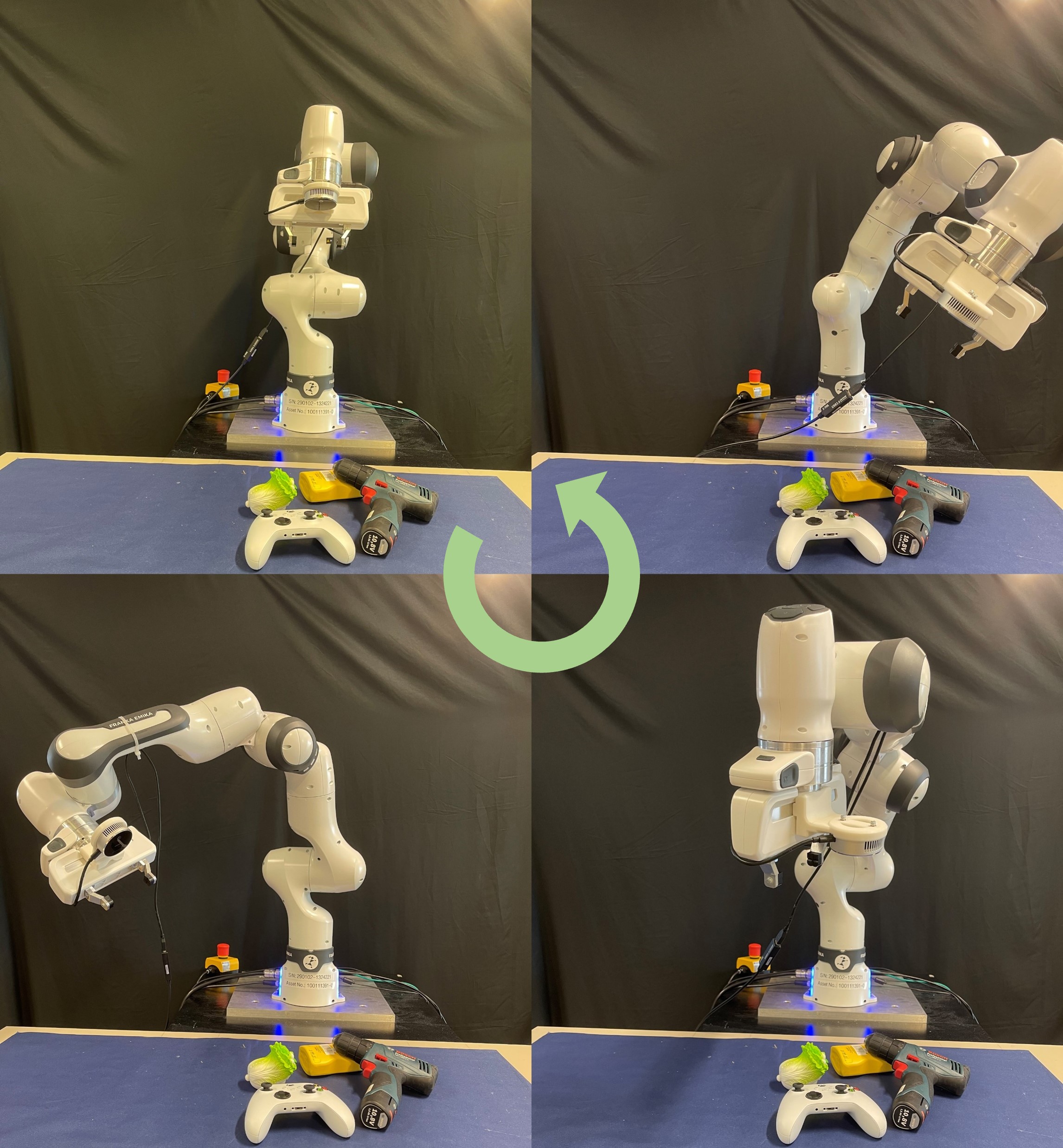}
	\caption{Configuration of real-world experiment.}
	\label{scan}	
\vspace{-4mm}
\end{figure}

 \begin{table*}[!t]
	\captionsetup{justification=centering}
	\renewcommand\arraystretch{1.0}
    \setlength\tabcolsep{10pt}
	\begin{center}
		\caption{A comparison with SOTA methods on GraspNet-1Billion dataset.}
		\label{results}
		\begin{tabular}{p{3.6cm}<{\centering}|p{0.6cm}<{\centering}|p{0.6cm}<{\centering}|p{0.6cm}<{\centering}|p{0.6cm}<{\centering}|p{0.6cm}<{\centering}|p{0.6cm}<{\centering}|p{0.6cm}<{\centering}|p{0.6cm}<{\centering}|p{0.6cm}}
			\hline
			\hline
			\multirow{2}*{Model} & \multicolumn{3}{c|}{Seen} & \multicolumn{3}{c|}{Similar} & \multicolumn{3}{c}{Novel} \\
            \cline{2-10}
			& \multicolumn{1}{c}{\textbf{AP}} & \multicolumn{1}{c}{$\textbf{AP}_\textit{0.8}$} & \multicolumn{1}{c|}{$\textbf{AP}_\textit{0.4}$} & \multicolumn{1}{c}{\textbf{AP}} & \multicolumn{1}{c}{$\textbf{AP}_\textit{0.8}$} & \multicolumn{1}{c|}{$\textbf{AP}_\textit{0.4}$} & \multicolumn{1}{c}{$\textbf{AP}$} & \multicolumn{1}{c}{$\textbf{AP}_\textit{0.8}$} & \multicolumn{1}{c}{$\textbf{AP}_\textit{0.4}$} \\
            \hline
            PointNet GPD~\cite{liang2019pointnetgpd} & 25.96 & 33.01 & 15.37 & 22.68 & 29.15 & 10.76 & 9.23 & 9.89 & 2.74 \\
            GraspNet Baseline~\cite{fang2020graspnet} & 27.56 & 33.43 & 16.95 & 26.11 & 34.18 & 14.23 & 10.55 & 11.25 & 3.98 \\
            Li \textit{et al.}~\cite{li2021simultaneous} & 36.55 & 47.22 & 19.24 & 28.36 & 36.11 & 10.85 & 14.01 & 16.56 & 4.82 \\
            RGB Matters~\cite{gou2021RGB} & 27.98 & 33.47 & 17.75 & 27.23 & 36.34 & 15.60 & 12.25 & 12.45 & 5.62 \\
            SB Baseline~\cite{Ma_2022_CoRL} & 58.95 & 68.18 & 54.88 & 52.97 & 63.24 & 46.99 & 22.63 & 28.53 & 12.00 \\
            \textbf{YOSO (Ours)} & \textbf{61.22} & \textbf{71.40} & \textbf{55.79} & \textbf{59.21} & \textbf{70.94} & \textbf{52.52} & \textbf{25.60} & \textbf{32.43} & \textbf{13.43} \\
			\hline
            \hline
		\end{tabular}
	\end{center}
\end{table*}

\section{EXPERIMENTS}

\subsection{Benchmark and Metric}
GraspNet-1Billion~\cite{fang2020graspnet} is a well-acknowledged dataset in 6-DoF robotic grasping, which includes 190 cluttered scenes captured in the real world by Kinect/Realsense camera in 256 views. Images are obtained by moving a robotic arm along predefined paths, covering 256 unique viewpoints on a quarter sphere. 
To assess the quality of grasps generated in complex, cluttered environments, we employ a precision-based evaluation metric. Following the methodology outlined in \cite{fang2020graspnet}, $\textbf{AP}_{\mu}$  is computed to signify the average \textit{Precision@k} across a range of k values spanning from 1 to 50 with friction $\mu$, and $\textbf{AP}$ is obtained by the average of $\textbf{AP}_{\mu}$, where $\mu$ varies from 0.2 to 1.2.



\subsection{Implementation Details}
To demonstrate that the reconstructed scene point cloud generated by our pipeline has a more informative understanding of objects in the scene to overcome occlusion from a single view and further facilitates the grasp generation process, we conduct experiments on the GraspNet-1Billion dataset by reconstructing meshes of all objects in a scene throughout all 256 views and estimate the object pose of each object in each view. Then grasp poses are generated on the reconstructed scene with the pre-trained model from Scale-balanced GraspNet~\cite{Ma_2022_CoRL}. However, some objects are severely occluded or even fully invisible in the first few or even half of the frames in a scene, which violates one of the assumptions of our pipeline that the object should be visible in the first few frames, it is inevitable that few of objects can not be tracked or reconstructed with BundleSDF \cite{wen2023bundlesdf}. In experiments, we manually remove objects with visually unacceptable mesh reconstruction of pose tracking results.

In our real-world experiment, an Intel RealSense L515 RGB LiDAR depth camera is mounted on a Franka Emika Panda robot arm as shown in Fig. \ref{scan}. The robotic arm scans the workspace by moving the end effector along a motion trajectory that covers the hemispherical area above the workspace. All three modules run on an NVIDIA RTX3090 GPU.


\section{RESULTS AND ANALYSIS}

\begin{figure*}[t]
	\centering
	\includegraphics[width=\linewidth]{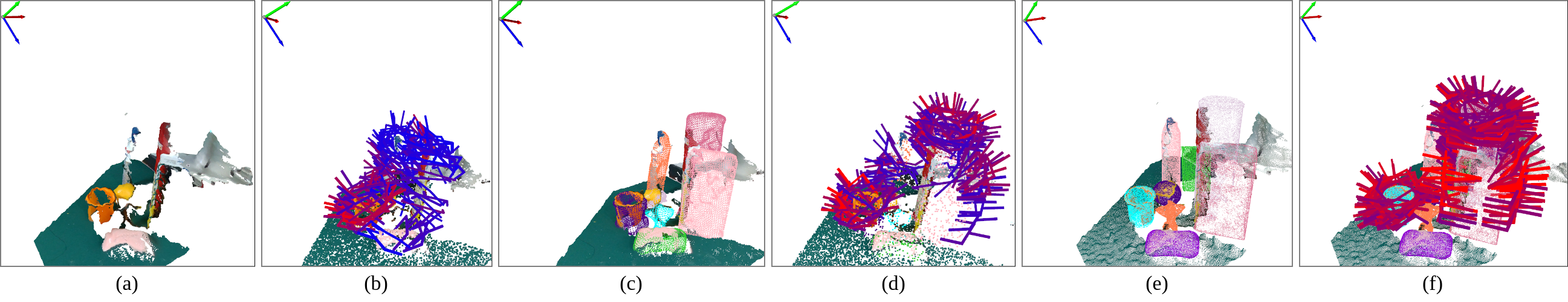}
	\caption{Qualitative comparison of grasp prediction with partial point cloud and reconstructed scene on GraspNet-1Billion dataset. Color varies from red to blue to represent the grasp quality from high to low. (a). Partial point cloud back-projected from depth image. (b). Grasps that are generated on a partial point cloud. (c). Reconstructed scene from YOSO pipeline. (d). Grasps that are generated on the reconstructed scene. (e). Complete scene-level point cloud. (f). Grasps generated on the complete scene-level point cloud.}
	\label{qualitative}	
\vspace{-4mm}
\end{figure*}

\subsection{Comparison with State-of-the-art Methods}
We present a comprehensive analysis of the experimental outcomes on the GraspNet-1Billion benchmark, spanning all three distinct test sets (seen, similar, and novel), as shown in Table \ref{results}. Notably, evaluation result reveals that our method yields substantial improvements in grasp generation accuracy when compared to current SOTA models that take partial point clouds as input. Instead, missing parts of objects are recovered through our YOSO pipeline, which provides reconstructed point clouds to the grasp pose prediction network and significantly boosts the accuracy of grasp generation.



\subsection{Effect of Scene Reconstruction on Grasp Generation Accuracy}
For robotic grasping and manipulation tasks, having a complete and detailed representation of the environment and objects within it is vital. This entails the reconstruction of occluded parts of objects that are absent in a single-view partial point cloud. The YOSO pipeline comprises two pivotal stages: \textbf{Stage I}, which focuses on generating detailed object meshes, and \textbf{Stage II}, which is dedicated to tracking object poses to seamlessly reintegrate the generated meshes into the overall scene. The quality of the perfect scene reconstruction hinges on the flawless execution of both pose tracking and mesh generation processes.

To assess the potential of our proposed pipeline, we enhance the GraspNet-1Billion dataset by creating a detailed visible point cloud for each scene. We compile these visible points from different perspectives to form a complete scene-level point cloud, using ground truth segmentation masks from 256 unique views and corresponding object CAD models.

We then use the pre-trained Scale-Balanced model \cite{Ma_2022_CoRL} to systematically assess grasp generation accuracy across three different input conditions: partial point clouds, scene point clouds reconstructed via YOSO, and our complete scene reconstruction from ground truth data, noted as Fully Visible. The results, presented in Table \ref{GT_compare}, suggest that providing more detailed input point clouds can potentially improve the performance of grasp prediction networks. However, while YOSO provides substantial improvements, as shown in Table \ref{results}, there is still a notable gap when compared to the outcomes achieved with complete scene reconstructions.

Our pipeline's modular design allows for future improvements, which may include enhancing the \textit{Video-segmentation Module} and improving the \textit{Object Pose Tracker and Mesh Generator} components with more sophisticated algorithms, all while maintaining real-time processing capabilities.

In our qualitative evaluation using the GraspNet-1Billion dataset, we examine the impact of different levels of scene reconstruction detail on grasp generation, as illustrated in Fig. \ref{qualitative}. The analysis indicates that as the quality of scene reconstruction increases, the diversity and accuracy of the generated grasps improve correspondingly.

 \begin{table}[!t]
	\captionsetup{justification=centering}
	\renewcommand\arraystretch{1.2}
    \setlength\tabcolsep{7pt}
	\begin{center}
		\caption{Comparison of grasp generation accuracy between different qualities of input point cloud. Partial represents partial point cloud back-projected from a depth image. YOSO (Ours) represents scene point cloud reconstructed by our YOSO pipeline. Fully Visible represents scene-level fully visible point cloud, which is regarded as a perfect reconstructed scene.}
		\label{GT_compare}
		\begin{tabular}{|p{2.6cm}<{\centering}|p{1.2cm}<{\centering}|p{1.2cm}<{\centering}|p{1.2cm}<{\centering}|}
        \hline
		{\diagbox{\textbf{Input PC}}{\textbf{AP}}} & \textbf{Seen} & \textbf{Similar} & \textbf{Novel} \\
        \hline
        Partial & 58.95& 52.97 & 22.63\\
			\hline
       \textbf{YOSO (Ours)} & \textbf{61.22}& \textbf{59.21} & \textbf{25.60}\\
			\hline
        Fully Visible & 69.76& 65.58 & 29.73\\
			\hline
		\end{tabular}
	\end{center}
\end{table}

\subsection{Inference Time Analysis}
In our proposed pipeline, a scene is only scanned once in \textbf{Stage I} to record RGB-D video for target object registration, which can be regarded as preparation phase. Therefore, we focus more on analyzing the inference time of each module during testing phase (\textbf{Stage II}), including 
Sec. \ref{sec:IIIA} \textit{Video-segmentation Module} (XMem), Sec. \ref{sec:IIIB} \textit{Object Pose Tracker} (BundleSDF) and Sec. \ref{sec:IIIC} \textit{Grasp Pose Predictor} (Scale-Balanced GraspNet), as shown in Table \ref{speed_test}. For \textit{Video-segmentation Module} and and \textit{Grasp Pose Predictor}, they takes input as a whole for prediction. For \textit{Object Pose Tracker}, the incorporated BundleSDF can only process one instance each time. When handling a single object, the overall inference of our pipeline amounts to 359.3 ms, in which scene reconstruction (combining XMem and BundleSDF) occupying merely 109.3 ms. Therefore the scene reconstruction runs at  9.2 frames per second (FPS) and the whole YOSO pipeline runs at 2.8 FPS. For scenarios with $M$ target objects, overall inference becomes $76 * M + 283.3$ ms if object tracking is performed sequentially, which is still faster than 1 FPS if $M < 10$. Alternatively, object pose tracking can be performed in multiple threads, leading to similar overall inference as single-object situation.


 \begin{table}[!t]
	\captionsetup{justification=centering}
	\renewcommand\arraystretch{1.2}
    \setlength\tabcolsep{8.5pt}
	\begin{center}
		\caption{A breakdown of the time required to execute each module in the pipeline.}
		\label{speed_test}
		\begin{tabular}{p{0.6cm}<{\centering}|p{0.8cm}<{\centering}|p{1.3cm}<{\centering}|p{1.8cm}<{\centering}|p{0.8cm}<{\centering}}
		\hline
        \hline
		& \multirow{2}*{\textbf{XMem}} & \multirow{2}*{\textbf{BundleSDF}} & \textbf{Scale-balanced GraspNet} & \multirow{2}*{\textbf{Total}} \\
        \hline
        \textbf{Time} & 33.3ms& 76ms & 250ms & 359.3ms\\
			\hline
            \hline
		\end{tabular}
	\end{center}
\end{table}

\section{CONCLUSIONS}

In this research work, we introduce our innovative dynamic scene reconstruction pipeline, YOSO, tailored for 6-DoF robotic grasping of novel objects. This pipeline streamlines the process by capturing the workspace in a single scan, generating object meshes, and storing essential key features for each novel object. During the testing phase, it continues to track the object's pose, seamlessly integrating the generated mesh back into the scene. By taking the reconstructed scene point cloud as input, accuracy and diversity of grasping strategies predicted by our incorporated grasp generation algorithm are significantly enhanced. Our pipeline reconstruct the scene at near real-time while providing more informative scene point cloud to the grasp generation algorithms, which is more applicable in real-world robotic grasping. Our comprehensive evaluation, conducted against the perfect scene visible point cloud data, demonstrates the promising potential for enhancing scene reconstruction quality, thereby achieving higher accuracy in grasp generation.





\section*{ACKNOWLEDGMENT}
This research is supported by the Agency for Science, Technology and Research (A*STAR) under its AME Programmatic Funding Scheme (Project \#A18A2b0046).


\bibliographystyle{IEEEtran}
\bibliography{root}

\end{document}